\pdfoutput=1

\documentclass[11pt]{article}

\usepackage{EMNLP2024}

\usepackage{times}
\usepackage{latexsym}

\usepackage{booktabs}
\usepackage{amsmath}
\usepackage{tcolorbox}
\usepackage[T1]{fontenc}

\usepackage[utf8]{inputenc}

\usepackage{microtype}

\usepackage{enumitem}

\usepackage[scaled=.8]{beramono}

\usepackage{soul}

\usepackage{lscape} 
\usepackage{rotating}

\title{Back to School: Translation Using Grammar Books}

\author{Jonathan Hus\textsuperscript{$\alpha$} \ \ Antonios Anastasopoulos\textsuperscript{$\alpha,\beta$} \\
  \textsuperscript{$\alpha$}Department of Computer Science, George Mason University, VA, USA \\
  \textsuperscript{$\beta$}Archimedes AI Unit, Athena Research Center, Athens, Greece\\
  \texttt{\{jhus,antonis\}@gmu.edu} }

\begin{document}
\maketitle
\begin{abstract}
Machine translation systems for high resource languages perform exceptionally well and produce high quality translations. Unfortunately, the vast majority of languages lack the quantity of parallel sentences needed to train such systems. These under-represented languages are not entirely without resources, as bilingual dictionaries and grammar books may be available as linguistic reference material. With current large language models (LLMs) supporting near book-length contexts, we can %
use the available material to ensure advancements are shared among all of the world's languages. In this paper, we %
use dictionaries and grammar books
to improve machine translation. We evaluate %
on 16 typologically diverse low-resource languages, showing encouraging improvements.%
\footnote{Code and data to reproduce our experiments are here: \url{https://github.com/jonathanhus/back-to-school}.}

\end{abstract}

\section{Introduction}

Machine translation systems have progressed remarkably, but they require massive amounts of parallel sentences \cite{bapna2022building}.
More recently, instruction-tuned large language models (LLMs) have also proven capable of performing machine translation. However, their performance is best when translating among high-resource languages that were most likely seen during training. 
Current transformer-based state-of-the-art large language models and mutlilingual translation models are trained on huge web-scraped corpora, with data in the order of trillions of tokens. 

While the web is a vast resource of good training data,\footnote{assuming aggressive filtering techniques} the web is also mainly comprised of just a handful of languages. 
There are an estimated 7000 languages in the world, but just 10 languages cover 84\% of the web content, with English covering more than 50\%. Therefore, low-resource languages are not well-represented in the training data for the large language models~\cite{joshi-etal-2020-state}, leading to systematic performance disparities across languages~\cite{blasi-etal-2022-systematic}. More importantly, language translation systems rely on a large number of parallel sentences, providing examples of sentences in the source and target languages. Therefore, the sheer magnitude of data that current translation systems require is simply not available for low resource languages. Given these constraints, the compelling question is: how can we create well-performing translation systems for low resource languages?

One approach to enabling machine translation for low-resource languages is to collect many parallel sentences. However, this is laborious, expensive, and time-consuming, requiring the skills of linguists and native speakers. Another approach would be to incorporate language reference material into the translation process of the LLM. The advantage of this approach is that a good number of dictionaries and grammar books have been created over decades (and longer) and require little additional effort to use them. 

In this work, we push the frontier using the latter approach to improve on the ability of LLMs to perform machine translation of low-resource languages by utilizing available linguistic reference materials. We incorporate dictionaries, grammar books, and a small number of parallel sentences into the prompt of a state-of-the-art LLM. We evaluate on 16 typologically diverse low-resource languages, performing analyses using different combinations of reference materials. %

\section{Related Work}
While tens of high-resource languages have enjoyed the recent advances in machine translation, many of the world's 7000+ languages have been unable to partake in the success. %

The current state of the art in multilingual and low-resource translation is the No Language Left Behind model~\cite{nllb2022}, relying on a mined and curated corpus of parallel sentences for 200 languages, including many low-resource ones. A large multilingual encoder-decoder translation model was then trained on this data to create a machine translation system for these languages.

On the other end of the spectrum, \citet{tanzer2023mtob} incorporated dictionaries, sentences, and grammar books to perform machine translation in a zero-shot setting, i.e., in a language without \textit{any} other data available, akin to how a documentary linguist or any second-language learner might learn a new language ("Machine Translation from One Book (MTOB)"). This paper inspired our own work, as it provides a framework for using LLMs to perform translation of resource-scarce languages. However, they were limited in the size of the context for the models they chose, and therefore, were only able to extract smaller chunks of the grammar book for inclusion. Here, we explore this paradigm in a much larger scale, with 15 more languages, performing additional necessary analyses.

Last, \citet{zhang2024hire} explored a similar path utilizing grammar books. They were also limited by the size of the model context, but they additionally used a morphological analyzer on the grammar books to extract linguistic features to assist in translation. Such tools are unfortunately unavailable for all languages, making this approach not feasible for scaling to thousands of languages.

\section{Preliminaries and Problem Definition}
 A traditional neural MT system models $p_{\text{MT}}(\mathbf{y} | \mathbf{x})$, learned over source-target sentence pairs $\langle \mathbf{x},\mathbf{y}\rangle$. At inference time, given a new source sentence, we sample a high-probability output from the learned distributions.
 A SOTA LLM, however, is first pre-trained to model $p_{\text{LM}}(x)$ and then instruction-tuned on $p_{\text{LM-ins}}(\mathbf{y}|\pi)$ over prompt-target text pairs $\langle \mathbf{\pi},\mathbf{y}\rangle$ covering multiple downstream tasks (often including MT). At inference time, with a similar prompt we sample outputs from the final model. %

A translation prompt $\pi(\cdot)$ at a minimum needs to include the task definition $\mathrm{t}$ (e.g. \texttt{"Please translate the following sentence to French:"}) and the source sentence $\mathbf{x}$: $\pi(\mathbf{x},\mathrm{t})$. For learning to translate an entirely unseen language, \citet{tanzer2023mtob} crafted prompts $\pi(\mathbf{x}, \mathrm{t}, \mathrm{d}, \mathrm{s}, \mathrm{g})$ that additionally included: 
\begin{itemize}[noitemsep,nolistsep,leftmargin=*]
    \item word-level translations $\mathrm{d}$ obtained from a bilingual dictionary $\mathcal{D}$, selected for their similarity to the words of the given source sentence, 
    \item a few parallel sentence examples $\mathrm{s}$, selected from a small collection of parallel sentences $\mathcal{S}$ for their similarity to the given source sentence, and
    \item excerpts $\mathrm{g}$ from a grammar book $\mathcal{G}$, also selected for similarity to the source sentence using longest common substring distance.
\end{itemize}

\section{Experiments}

\paragraph{Languages} \ We focus on 16 largely under-served low-resource languages, chosen for geographical and typological diversity, as well as resource (dictionary, grammars) and evaluation data availability. Specifically, we work with: Chokwe, Chuvash, Dinka, Dogri, Gitksan, Guarani, Ilokano, Kabuverdianu, Kachin, Kalamang, Kimbundu, Latgalian, Minangkabau, Mizo, Natugu, and Wolof. We evaluate translation both into and out of English. %

\paragraph{Dictionaries} \ 
We obtain dictionaries from PanLex\footnote{\url{https://panlex.org}} for all our languages.  Note that, in cases where the number of words in the dictionary was less than 100 we do not include them in the prompt. The size of each dictionary is included in Appendix~\ref{app:resources}.

\paragraph{Parallel Sentences} \ 
For the parallel sentences that are part of the prompts as translation examples, we use the dev portion of the FLORES-200 dataset.\footnote{\url{https://github.com/openlanguagedata/flores}}
Gitksan and Natugu are not represented in FLORES and instead we use the data that \citet{zhang2024hire} provided.

\begin{table*}[t]
\centering
\begin{tabular}{l|c@{ }c@{ }c@{ }c@{ }c|c@{ }c@{ }c@{ }c@{ }c}
\toprule
 & \multicolumn{5}{c|}{English$\rightarrow$X} & \multicolumn{5}{c}{X$\rightarrow$English} \\
\textbf{Language} & \textbf{Baseline} & \textbf{W} & \textbf{W+S} & \textbf{W+S+G} & \textbf{NLLB} & \textbf{Baseline} & \textbf{W} & \textbf{W+S} & \textbf{W+S+G} & \textbf{NLLB} \\
\midrule
\multicolumn{7}{@{}l}{\small{\textbf{Languages supported by NLLB with some online presence}}}\\
Chokwe & 12.3 & - & \textbf{21.0*} & 16.9 & \underline{24.3}& 22.8& - & \textbf{27.3*}& 25.8 & \underline{30.8}\\
Dinka & 8.8 & - & \textbf{16.3*} & 11.1 & \underline{24.2} &  20.7& -& \textbf{25.0*}& 23.0 & \underline{31.2}\\
Guarani & \textbf{29.4} & 20.6 & 29.1 & 29.0 & \underline{36.9} & \textbf{43.4*}& 41.7& 42.3& 41.7 & \underline{48.4}\\
Ilokano & 43.1 & 37.6 & \textbf{45.1*} & 43.8 & \underline{53.3} & \textbf{53.9*}& 52.1& 52.5& 53.6 & \underline{62.1}\\
Kabuverdianu & 39.0 & 29.8 & \underline{\textbf{55.9}}\textbf{*} & 47.2 & 42.8 & \underline{\textbf{69.3}}\textbf{*}& 66.9& 68.3& 68.4 & 68.4 \\
Kachin & 12.5 & - & \textbf{27.7*} & 21.2  & \underline{37.5} & 22.5& -&\textbf{25.2*}& 23.8 & \underline{41.6}\\
Kimbundu & 11.6 & - & \underline{\textbf{26.2}}\textbf{*} & 14.4 & 24.9 & 19.3& -& 24.3& \textbf{25.0*} & \underline{33.9} \\
Latgalian & 26.0 & 21.0 & \textbf{37.6*} & 31.1 & \underline{53.6} & 49.8& 41.1& 48.5& \textbf{50.3} & \underline{63.4}\\
Minangkabau & 42.0 & 28.1 & \textbf{47.0*} & 44.3 & \underline{52.4} & \textbf{55.1*}& 43.9& 51.9& 54.0& \underline{62.5} \\
Mizo & 30.4 & 29.7 & \textbf{32.2} & 30.3 & \underline{38} &  \textbf{36.6*}& 35.0& 35.6& 36.2& \underline{41.4} \\
Wolof & 23.2 & 15.0 & 25.6 & \textbf{26.0} & \underline{29.7} & \textbf{36.4*}& 29.6& 31.3& 35.8 & \underline{41.2}\\
\multicolumn{7}{@{}l}{\small{\textbf{Languages \textit{not }supported by NLLB with minimal online presence}}}\\
Chuvash	& 2.6 & 13.7 & \textbf{19.0*} & 16.0 & -- & 25.4& 23.4& 24.2& \textbf{26.8*} & -- \\
Dogri & 5.9 & - & \textbf{34.3*} & 24.9 & -- &  51.2& -& \textbf{52.4*}& 52.0 & --\\
Gitksan & 7.8 & - & 13.3 & \textbf{15.9*} & -- & 14.0& -& 24.4& \textbf{24.6} & --\\
Kalamang & 5.1 & 27.1 & \textbf{41.9*} & 37.3  & --  & 11.3& 18.7& 27.6& \textbf{34.8*} & --\\
Natugu & 6.8 & 4.5 & 12.0 & \textbf{17.0*} & -- & 13.2& 6.8& 9.9& \textbf{23.7*} &--\\
\midrule
\multicolumn{1}{l}{\textbf{System Average:}} & 19.2 & 22.7 & 30.3 & 26.7 & 38.0$^\dagger$ & 34.1 & 35.9 & 35.7 & 37.5 & 47.7$^\dagger$\\
\multicolumn{1}{l}{\textbf{System Wins:}} & 1 & 0 & 12 & 3 & (9/11)$^\dagger$ & 6 & 0 & 4 & 6 & (10/11)$^\dagger$\\
\bottomrule
\end{tabular}
\vspace{-.5em}
\caption{Collective Table of Results (chrF++ scores). The combination of reference material that led to the best score is \textbf{bolded}. We also compare to NLLB, with the best score underlined. An asterisk ('*') indicates that the difference between our best system and the others is statistically significant. System wins counts the best combination of reference material among our systems (NLLB excluded). $^\dagger$: NLLB only supports 11 of our languages.}
\label{tab:all-results}
\vspace{-0.5em}
\end{table*}

\paragraph{Grammar Books} \ 
The DReaM corpus \cite{virk-etal-2020-dream} contains digitized versions of thousands of linguistic documents, including grammar books and sketches, for many languages. The source of these documents is often in paper format, and due to the scanning/OCR quality, the digitized versions often contain scanning artifacts. 
We select one grammar document for each of our languages (concrete details in Appendix~\ref{app:resources}). 
We perform slight manual cleanup to remove some items (e.g., scanning artifacts, table of contents) and to ensure that the grammar would fit in the LLM's context size. 

\paragraph{Evaluation} \ We use the devtest portion of FLORES-200 as our evaluation set. For Gitksan and Natugu, we use the test sets from the SIGMORPHON 2023 shared task \cite{ginn-etal-2023-findings}. We report chrF++ scores~\cite{popovic-2017-chrf} for both language directions. 

\subsection{Model}
We use the GPT-4-turbo model for our experiments. In addition to being the latest offering from OpenAI (and presumably its most capable, at the time of writing), it has an input context size of 128K. This large context enables book-length text to be included in the prompt. The grammar books we use range from tens of pages to a couple hundred pages in length, which equates to roughly 40K to 120K tokens. Models with such capacity have only recently been made available, which affords us the opportunity to use full-length grammar books as opposed to smaller heuristically-selected excerpts. %

\paragraph{Prompt Format} \ 
Our prompts largely follow the MTOB framework, using complete prompts $\pi(\mathbf{x}, \mathrm{t}, \mathrm{d}, \mathrm{s}, \mathrm{g})$ with task instructions and source sentence (provided in the prompt beginning and repeated at the end), as well as word pairs from the dictionary, example sentences, and the language's grammar. We perform ablations removing components from the prompt to establish their contributions, e.g. repeating all experiments without incorporating the grammar book, i.e. using $\pi(\mathbf{x}, \mathrm{t}, \mathrm{d}, \mathrm{s})$.
We provide specific details as well as an example prompt in Appendix~\ref{app:prompt}. %

\section{Results}
Table~\ref{tab:all-results} shows the results for the experiments. We report results on both translation directions, with different prompt configurations as discussed above. %
We report two comparison models: Baseline corresponds to 0-shot LLM translation performance i.e., only with prompt $\pi(\mathbf{x}, \mathrm{t})$, and the "skyline" performance of NLLB, the current SOTA multilingual MT model. We also report results by adding words (\textbf{W}: $\pi(\mathbf{x}, \mathrm{t}, \mathrm{d})$), sentences (\textbf{W+S}: $\pi(\mathbf{x}, \mathrm{t}, \mathrm{d}, \mathrm{s}$), and grammars (\textbf{W+S+G}: $\pi(\mathbf{x}, \mathrm{t}, \mathrm{d}, \mathrm{s}, \mathrm{g})$) to the prompt.

For each language and direction, we have four systems that we compare. We compute all evaluation metrics using SacreBLEU~\footnote{nrefs:1|case:mixed|eff:yes|nc:6|nw:2|space:no|version:2.4.0}~\cite{post-2018-call} and we also report statistical significance using paired bootstrap resampling, comparing our best-performing system to the other systems. In most cases, we find that the difference is statistically significant, indicating that the translation performance is dependent on the selected prompt content.

\subsection{Comparison to SOTA MT}
We compare the best results we achieved with the chrF++ scores from NLLB, for the languages supported by NLLB. Note that these are languages with at least some online presence. 
In general the NLLB scores were better, but there were a few instances where our approach outperformed NLLB. When going from English to a target language, including words and sentences in the prompt for Kabuverdianu and Kimbundu provided the best results. 
For Kabuverdianu, including the grammar book also surpassed the NLLB score. 
When translating Kabuverdianu into English, the baseline model (0-shot) with no reference material was best. Kabuverdianu, as a Portuguese-based Creole, has many similarities to Portuguese, a high resource language. This might explain this result and it could be reflective of GPT-4's %
capabilities. 

\subsection{Sentences or Grammar Books?}
The results of our experiments show that the inclusion of grammar books does not always lead to the best score (see bottom rows of Table~\ref{tab:all-results}). In fact, when translating from English, using only words and sentences yields the highest score for 12 of the languages. When translating into English, the combination of words, sentences, and grammar books had the highest score for six of the languages. However, including no reference material at all was the best approach for six languages as well.

To explore the reasons behind these results, we perform a linear regression that aims to predict the score of the $W$+$S$+$G$ combination given the baseline score and the following features:
\begin{itemize}[noitemsep,nolistsep,leftmargin=*]
    \item Number of words in the reference dictionary
    \item Number of sentences available in corpora as reported in OPUS \cite{4992de1b5fb34f3e9691772606b36edf}
    \footnote{\url{https://opus.nlpl.eu/}} 
    \item Perplexity of the grammar book 
    \item Length of the grammar book in tokens
\end{itemize}

\noindent The features regarding words and sentences correspond directly to data availability, with the assumption that more data is better. The grammar book features are proxies for the quality and the completeness of the documented grammar. For perplexity, we used a GPT-2 model and passed the grammar book as input to the model. LM perplexity is then measured using a sliding window strategy.

The $R^2$ values for these regressions are listed in Table~\ref{tab:feature-importance}. Put simply, the $R^2$ value denotes the quality of the model fit, and can help us determine the percentage of variance in the dependent variable (downstream performance, in our case) that can be explained by the independent variable.

We find that the number of dictionary words and the length of the grammar books have a positive influence on the score, while the perplexity has a negative impact. While this aligns with our expectations, a finding that is seemingly surprising is that the number of available sentences has a negative impact on the score compared to the baseline. 
This necessitates further research to actually confirm, but we suspect that this is because GPT-4 has already been pre-trained on data from these languages and, consequently, it can perform better on them. 
This is most pronounced when translating into English, where the top 5 languages (by number of sentences) all perform best under the baseline setting i.e., no additional reference material. All languages that are best translated using no reference material appear before all of the languages that are best translated using the combination of dictionaries, parallel sentences, and grammar books. This suggests that %
using grammars might be best suited to extremely low-resource languages with less than $10^3$ %
parallel sentences. 

\begin{figure}[t]
    \centering
    \includegraphics[width=\columnwidth]{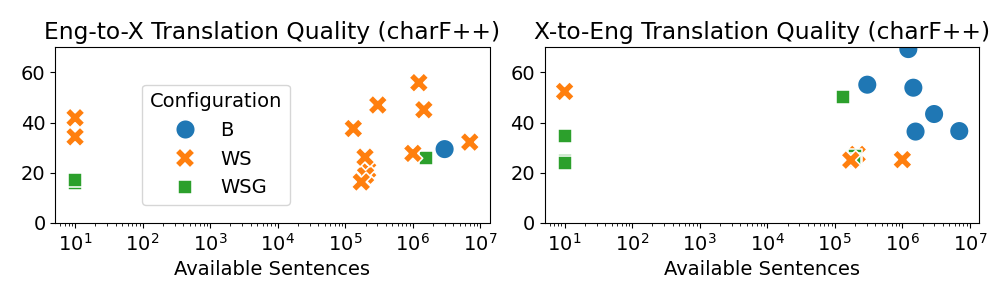}
    \vspace{-2em}
    \caption{Using grammars is particularly beneficial for extremely low-resource languages. Simple prompt-based MT (zero-shot) is best for high-resource ones.}
    \label{fig:enter-label}
\end{figure}

\begin{table}[t]
\centering
\small
\begin{tabular}{@{}l|c@{ }c|c@{ }c@{}}
\toprule
 & \multicolumn{2}{c}{eng $\rightarrow$ X} & \multicolumn{2}{c}{X $\rightarrow$ eng} \\
\textbf{Language} & \textbf{Add.} & \textbf{Single} & \textbf{Add} & \textbf{Single}\\
\midrule
Baseline & 0.643 & 0.643 & 0.849 & 0.849 \\
 + Words & 0.648 & 0.054 & 0.850 & 0.007\\
 + Sentences & 0.708 & 0.050 & 0.880 & 0.012 \\
 + Perplexity & 0.751 & 0.177 & 0.925 & 0.141 \\
 + Length & 0.755 & 0.062 & 0.927 & 0.115 \\
\bottomrule
\end{tabular}
\caption{\label{tab:feature-importance}
$R\textsuperscript{2}$ values for features explaining the $W$+$S$+$G$ chrF++ output. "Add.": incorporating the feature with the ones above. "Single": linear regression with only that feature as input.}
\end{table}

\section{Conclusion}
In this paper, we showed that utilizing reference material such as dictionaries and grammar books in the prompt of an LLM can improve the performance of machine translation for low-resource languages. We evaluated the performance on 16 languages and showed that the improvement is especially pronounced for languages that have minimal presence on the web. Our work shows that this approach has the potential to address the gap for extremely low-resource languages and identifies a concrete path for improving MT for more than~2,000 languages.

\section*{Limitations}
A primary contribution of this paper is the use of full-length grammar books in the input prompt in order to "teach" a model how to translate into a given language. However, there are some limitations with this approach. First, high quality grammar books are difficult to obtain for many languages. The DReaM corpus does an admirable job of curating and digitizing many linguistic references, but the output is not perfect. Multi-column text documents and tables lose information that is conveyed by the location of text relative to other text on the page. The LLMs, therefore, are most likely not taking full advantage of that information. Additionaly, scanning artifacts like headers and page numbers add unnecessary clutter to the reference material.

At the time of this writing, GPT-4-turbo was the only available model with the desired context length of 128K. Running the experiments using a set of models would indicate whether the reference material is improving translations or whether the model itself (and its associated training) is responsible for the performance.

The sizes of the bilingual dictionaries were inconsistent, with a handful having less than 20 words. We removed these low-volume dictionaries from our experiments. However, larger dictionaries of similar magnitudes would most likely improve the translations and would allow translation performance across the various languages to be better compared.

Finally, these experiments are not cheap. We estimate that all these experiments cost around \$15,000 USD using the standard pricing tier under the Azure Open AI Studio. This could significantly hinder the reproducibility of our results.  %

\section*{Ethics Statement}
We do not anticipate any ethical issues arising from our work.

\section*{Acknowledgements}
We are thankful to the reviewers and meta-reviewer for their constructive feedback.
This work was generously supported by the National Science Foundation under grant IIS-2327143.
It has also benefited from resources provided through the Microsoft Accelerate Foundation Models Research (AFMR) grant program.
This work was partially supported by resources provided by the Office of Research Computing at George Mason University (URL: \url{https://orc.gmu.edu}) and funded in part by grants from the National Science Foundation (Award Number 2018631).

\bibliography{anthology,custom}

\begin{thebibliography}{12}
\expandafter\ifx\csname natexlab\endcsname\relax\def\natexlab#1{#1}\fi

\bibitem[{Bapna et~al.(2022)Bapna, Caswell, Kreutzer, Firat, van Esch, Siddhant, Niu, Baljekar, Garcia, Macherey, Breiner, Axelrod, Riesa, Cao, Chen, Macherey, Krikun, Wang, Gutkin, Shah, Huang, Chen, Wu, and Hughes}]{bapna2022building}
Ankur Bapna, Isaac Caswell, Julia Kreutzer, Orhan Firat, Daan van Esch, Aditya Siddhant, Mengmeng Niu, Pallavi Baljekar, Xavier Garcia, Wolfgang Macherey, Theresa Breiner, Vera Axelrod, Jason Riesa, Yuan Cao, Mia~Xu Chen, Klaus Macherey, Maxim Krikun, Pidong Wang, Alexander Gutkin, Apurva Shah, Yanping Huang, Zhifeng Chen, Yonghui Wu, and Macduff Hughes. 2022.
\newblock \href {http://arxiv.org/abs/2205.03983} {Building machine translation systems for the next thousand languages}.

\bibitem[{Blasi et~al.(2022)Blasi, Anastasopoulos, and Neubig}]{blasi-etal-2022-systematic}
Damian Blasi, Antonios Anastasopoulos, and Graham Neubig. 2022.
\newblock \href {https://doi.org/10.18653/v1/2022.acl-long.376} {Systematic inequalities in language technology performance across the world{'}s languages}.
\newblock In \emph{Proceedings of the 60th Annual Meeting of the Association for Computational Linguistics (Volume 1: Long Papers)}, pages 5486--5505, Dublin, Ireland. Association for Computational Linguistics.

\bibitem[{Ginn et~al.(2023)Ginn, Moeller, Palmer, Stacey, Nicolai, Hulden, and Silfverberg}]{ginn-etal-2023-findings}
Michael Ginn, Sarah Moeller, Alexis Palmer, Anna Stacey, Garrett Nicolai, Mans Hulden, and Miikka Silfverberg. 2023.
\newblock \href {https://doi.org/10.18653/v1/2023.sigmorphon-1.20} {Findings of the {SIGMORPHON} 2023 shared task on interlinear glossing}.
\newblock In \emph{Proceedings of the 20th SIGMORPHON workshop on Computational Research in Phonetics, Phonology, and Morphology}, pages 186--201, Toronto, Canada. Association for Computational Linguistics.

\bibitem[{Joshi et~al.(2020)Joshi, Santy, Budhiraja, Bali, and Choudhury}]{joshi-etal-2020-state}
Pratik Joshi, Sebastin Santy, Amar Budhiraja, Kalika Bali, and Monojit Choudhury. 2020.
\newblock \href {https://doi.org/10.18653/v1/2020.acl-main.560} {The state and fate of linguistic diversity and inclusion in the {NLP} world}.
\newblock In \emph{Proceedings of the 58th Annual Meeting of the Association for Computational Linguistics}, pages 6282--6293, Online. Association for Computational Linguistics.

\bibitem[{Kamholz et~al.(2014)Kamholz, Pool, and Colowick}]{kamholz-etal-2014-panlex}
David Kamholz, Jonathan Pool, and Susan Colowick. 2014.
\newblock \href {http://www.lrec-conf.org/proceedings/lrec2014/pdf/1029_Paper.pdf} {{P}an{L}ex: Building a resource for panlingual lexical translation}.
\newblock In \emph{Proceedings of the Ninth International Conference on Language Resources and Evaluation ({LREC}'14)}, pages 3145--3150, Reykjavik, Iceland. European Language Resources Association (ELRA).

\bibitem[{{NLLB Team} et~al.(2022){NLLB Team}, Costa-jussà, Cross, Çelebi, Elbayad, Heafield, Heffernan, Kalbassi, Lam, Licht, Maillard, Sun, Wang, Wenzek, Youngblood, Akula, Barrault, Mejia-Gonzalez, Hansanti, Hoffman, Jarrett, Sadagopan, Rowe, Spruit, Tran, Andrews, Ayan, Bhosale, Edunov, Fan, Gao, Goswami, Guzmán, Koehn, Mourachko, Ropers, Saleem, Schwenk, and Wang}]{nllb2022}
{NLLB Team}, Marta~R. Costa-jussà, James Cross, Onur Çelebi, Maha Elbayad, Kenneth Heafield, Kevin Heffernan, Elahe Kalbassi, Janice Lam, Daniel Licht, Jean Maillard, Anna Sun, Skyler Wang, Guillaume Wenzek, Al~Youngblood, Bapi Akula, Loic Barrault, Gabriel Mejia-Gonzalez, Prangthip Hansanti, John Hoffman, Semarley Jarrett, Kaushik~Ram Sadagopan, Dirk Rowe, Shannon Spruit, Chau Tran, Pierre Andrews, Necip~Fazil Ayan, Shruti Bhosale, Sergey Edunov, Angela Fan, Cynthia Gao, Vedanuj Goswami, Francisco Guzmán, Philipp Koehn, Alexandre Mourachko, Christophe Ropers, Safiyyah Saleem, Holger Schwenk, and Jeff Wang. 2022.
\newblock No language left behind: Scaling human-centered machine translation.

\bibitem[{Popovi{\'c}(2017)}]{popovic-2017-chrf}
Maja Popovi{\'c}. 2017.
\newblock \href {https://doi.org/10.18653/v1/W17-4770} {chr{F}++: words helping character n-grams}.
\newblock In \emph{Proceedings of the Second Conference on Machine Translation}, pages 612--618, Copenhagen, Denmark. Association for Computational Linguistics.

\bibitem[{Post(2018)}]{post-2018-call}
Matt Post. 2018.
\newblock \href {https://www.aclweb.org/anthology/W18-6319} {A call for clarity in reporting {BLEU} scores}.
\newblock In \emph{Proceedings of the Third Conference on Machine Translation: Research Papers}, pages 186--191, Belgium, Brussels. Association for Computational Linguistics.

\bibitem[{Tanzer et~al.(2023)Tanzer, Suzgun, Visser, Jurafsky, and Melas-Kyriazi}]{tanzer2023mtob}
Garrett Tanzer, Mirac Suzgun, Eline Visser, Dan Jurafsky, and Luke Melas-Kyriazi. 2023.
\newblock A benchmark for learning to translate a new language from one grammar book.
\newblock In \emph{Arxiv}.

\bibitem[{Tiedemann(2009)}]{4992de1b5fb34f3e9691772606b36edf}
J{\"o}rg Tiedemann. 2009.
\newblock \emph{News from OPUS - A Collection of Multilingual Parallel Corpora with Tools and Interfaces}, volume~V, pages 237--248.

\bibitem[{Virk et~al.(2020)Virk, Hammarstr{\"o}m, Forsberg, and Wichmann}]{virk-etal-2020-dream}
Shafqat~Mumtaz Virk, Harald Hammarstr{\"o}m, Markus Forsberg, and S{\o}ren Wichmann. 2020.
\newblock \href {https://aclanthology.org/2020.lrec-1.110} {The {DR}ea{M} corpus: A multilingual annotated corpus of grammars for the world{'}s languages}.
\newblock In \emph{Proceedings of the Twelfth Language Resources and Evaluation Conference}, pages 878--884, Marseille, France. European Language Resources Association.

\bibitem[{Zhang et~al.(2024)Zhang, Choi, Song, He, Wang, and Li}]{zhang2024hire}
Kexun Zhang, Yee~Man Choi, Zhenqiao Song, Taiqi He, William~Yang Wang, and Lei Li. 2024.
\newblock \href {http://arxiv.org/abs/2402.18025} {Hire a linguist!: Learning endangered languages with in-context linguistic descriptions}.

\end{thebibliography}
\bibliographystyle{acl_natbib}

\pagebreak

\appendix

\section{Additional Experimental Results}
\label{app:other-results}
Table \ref{best-scores-ref} shows the best performing system for each language and direction, sorted in descending order by number of available sentences as reported by OPUS. 

Table \ref{tab:bootstrap-ek} and Table \ref{tab:bootstrap-ke} show the results from our paired significance tests. The best performing system for a given language and direction is compared to each of the other systems, with statistically significant differences indicated with an asterisk. 

The main paper uses chrF++ scores to evaluate translations, which is the metric used by NLLB. We also calculate BLEU scores for all of our experiments, which are provided in Table \ref{tab:bleu-results}.

\begin{table}
\centering
\begin{tabular}{@{}l|c@{ }c@{ }c@{ }}
\toprule
\textbf{Language} & \textbf{\# Sentences} & \textbf{eng $\rightarrow$ X} & \textbf{X $\rightarrow$ eng} \\
\midrule
mizo & 6979898 & WS & B \\
guarani & 2959865 & B & B \\
wolof & 1572603 & WSG &B \\
ilokano & 1458586 & WS &B \\
kabuverdianu & 1229409 & WS &B \\
kachin & 1003100 & WS & WS \\
minangkabau & 303354 & WS & B \\
chokwe & 214973 & WS & WS \\
chuvash & 200001 & WS & WSG \\
kimbundu & 196240 & WS & WSG \\
dinka & 172589 & WS & WS \\
latgalian & 131709 & WS & WSG \\
dogri & 0 & WS & WS \\
gitksan & 0 & WSG & WSG \\
kalamang & 0 & WS & WSG \\
natugu & 0 & WSG & WSG\\
\bottomrule
\end{tabular}
\caption{
Combination of reference material that led to the best score for each language, where B=baseline, W=words, WS=Words and Sentences, and WSG=Words, Sentences, and Grammar Book. Number of sentences is the total number of sentences as reported by OPUS.}
\label{best-scores-ref}
\end{table}

\section{Resources}
\label{app:resources}
For our experiments, we gathered dictionaries, parallel sentences, and grammar books to use in the prompts. Dictionaries were obtained from PanLex \cite{kamholz-etal-2014-panlex} and converted into the format required by the code. The dictionary used in MTOB included part of speech tags for each word, which is unavailable in PanLex. Therefore, we did not include this feature in our dictionaries. The sizes of the dictionaries are shown in Table \ref{tab:words-sentences-stats}. Kalamang is not available in PanLex, and we instead used the version from the MTOB paper.

For sentences, we used the FLORES dataset, originally released by Meta as FLORES-200\footnote{\url{https://github.com/facebookresearch/flores/blob/main/flores200/README.md}} and now maintained by the Open Language Data Initiative (OLDI) as FLORES+\footnote{\url{https://github.com/openlanguagedata/flores}}. For each language in the dataset, the dev split has 997 sentences and the devtest split has 1012 sentences. We used dev sentences as sample sentences in the prompts, while devtest sentences are used as translation tasks for our system on which performance was measured. For Dogri and Chuvash only the dev split is available. We therefore randomly split the dev split into dev and devtest with 497 and 500 sentences, respectively. Gitksan and Natugu are not represented in FLORES and we obtain sentences from the SIGMORPHON 2023 Shared Task on Interlinear Glossing,\footnote{\url{https://github.com/sigmorphon/2023glossingST}} which has dev, train, and test splits. These were combined to form dev and devtest splits. For Kalamang, the train and test splits as provided in the original paper were used unaltered. Table \ref{tab:words-sentences-stats} lists the sizes of the train and test splits for each of the languages.

Grammar books were obtained from the DReaM corpus, which contains digitized versions of numerous linguistic reference materials. When selecting the specific grammar book or sketch to use for each language, we searched for documents that provided a well-rounded description, appeared to have been well-processed by optical character recognition, and would fit within the context of GPT-4. For each document we performed limited formatting, such as removing the table of contents, in order to reduce the token count. Table \ref{tab:grammar-book-stats} lists the source documents used for the grammar books as well as the number of tokens for each document. Perplexity was measured using a GPT-2 model in order to provide a coarse assessment of the quality of the document. For Kalamang, we used the grammar book provided in MTOB. Specifically, we use the "long" version, which is a manually curated subset of Visser's grammar, that they tested on a Claude 2 model.

The authors of MTOB and the maintainers of FLORES+ explicitly request that this reference data, and the parallel sentences in particular, are not publicly hosted as plain text. This is to ensure that the resources are not web-scraped where they could potentially be included in the training data of future models, which would taint results of MT tests. In accordance with their requests, and with the same spirit in mind, we have password encrypted all reference material that we have posted and request that any users of our data do the same.

\vfill
\pagebreak

\begin{sidewaystable}
\centering
\footnotesize
\begin{tabular}{@{}l|c@{ }c@{ }c@{ }c@{}}
\toprule
\textbf{Language} & \textbf{Baseline} & \textbf{W} & \textbf{W+S} & \textbf{W+S+G}\\
\midrule
Chokwe&12.7 (12.7 +/- 0.2) p = 0.0010*&NA&21.9 (21.9 +/- 2.0)&17.9 (17.9 +/- 2.1) p = 0.0040*\\
Chuvash&4.2 (4.1 +/- 1.1) p = 0.0010*&13.5 (13.6 +/- 1.8) p = 0.0010*&19.2 (19.2 +/- 1.9)&17.2 (17.2 +/- 1.1) p = 0.0050*\\
Dinka&8.8 (8.8 +/- 0.2) p = 0.0010*&NA&17.5 (17.6 +/- 2.5)&11.1 (11.2 +/- 2.2) p = 0.0010*\\
Dogri&7.7 (7.7 +/- 2.6) p = 0.0010*&NA&34.2 (34.2 +/- 2.8)&25.0 (25.0 +/- 2.6) p = 0.0010*\\
Gitksan&8.5 (8.5 +/- 0.7) p = 0.0010*&NA&14.1 (14.2 +/- 2.2) p = 0.0010*&18.0 (18.0 +/- 1.3)\\
Guarani&30.1 (30.1 +/- 0.7)&20.9 (20.9 +/- 0.8) p = 0.0010*&30.0 (30.0 +/- 0.7) p = 0.3207&29.8 (29.8 +/- 0.7) p = 0.1279\\
Ilokano&43.9 (43.9 +/- 0.6) p = 0.0010*&38.4 (38.4 +/- 0.6) p = 0.0010*&45.9 (45.9 +/- 1.3)&44.6 (44.6 +/- 1.2) p = 0.0400*\\
Kabuverdianu&40.4 (40.4 +/- 1.1) p = 0.0010*&30.1 (30.0 +/- 1.5) p = 0.0010*&56.4 (56.3 +/- 1.2)&47.7 (47.7 +/- 1.0) p = 0.0010*\\
Kachin&12.6 (12.6 +/- 0.7) p = 0.0010*&NA&27.7 (27.8 +/- 3.8)&21.0 (21.2 +/- 3.3) p = 0.0020*\\
Kalamang&5.8 (5.8 +/- 0.5) p = 0.0010*&29.2 (29.1 +/- 4.1) p = 0.0010*&42.9 (43.0 +/- 4.8)&38.5 (38.5 +/- 5.3) p = 0.0150*\\
Kimbundu&11.9 (11.9 +/- 0.1) p = 0.0010*&NA&26.8 (26.9 +/- 2.0)&15.6 (15.6 +/- 2.0) p = 0.0010*\\
Latgalian&28.4 (28.4 +/- 0.8) p = 0.0010*&22.4 (22.5 +/- 1.1) p = 0.0010*&39.7 (39.7 +/- 1.0)&33.1 (33.1 +/- 0.8) p = 0.0010*\\
Minangkabau&42.8 (42.8 +/- 0.9) p = 0.0010*&29.0 (29.0 +/- 1.7) p = 0.0010*&47.8 (47.8 +/- 1.3)&45.4 (45.4 +/- 1.1) p = 0.0010*\\
Mizo&32.5 (32.5 +/- 0.8) p = 0.1678&29.7 (29.7 +/- 0.9) p = 0.1399&31.2 (31.3 +/- 3.3)&30.4 (30.5 +/- 2.5) p = 0.1988\\
Natugu&7.1 (7.1 +/- 0.5) p = 0.0010*&4.8 (4.9 +/- 0.9) p = 0.0010*&13.0 (13.2 +/- 3.2) p = 0.0010*&18.7 (18.7 +/- 3.0)\\
Wolof&24.0 (24.0 +/- 0.8) p = 0.0010*&15.5 (15.5 +/- 0.7) p = 0.0010*&26.3 (26.3 +/- 1.4) p = 0.1069&27.1 (27.1 +/- 0.7)\\
\bottomrule
\end{tabular}
\caption{\label{tab:bootstrap-ek}
Paired bootstrap resampling for chrF++ scores for English-to-X translations. The best performing system is selected as the baseline and is compared to the other systems, using a 95\% confidence interval and a p-value of 0.05. Statistically significant differences are noted with an asterisk (*).}
\end{sidewaystable}

\begin{sidewaystable}
\centering
\footnotesize
\begin{tabular}{@{}l|c@{ }c@{ }c@{ }c@{}}
\toprule
\textbf{Language} & \textbf{Baseline} & \textbf{W} & \textbf{W+S} & \textbf{W+S+G}\\
\midrule
Chokwe&23.5 (23.5 +/- 0.9) p = 0.0010*&NA&28.1 (28.1 +/- 0.7)&26.5 (26.5 +/- 0.7) p = 0.0010*\\
Chuvash&24.0 (24.0 +/- 2.5) p = 0.0140*&21.8 (21.9 +/- 1.9) p = 0.0010*&22.9 (22.9 +/- 2.1) p = 0.0010*&26.6 (26.6 +/- 2.5)\\
Dinka&21.6 (21.6 +/- 1.2) p = 0.0010*&NA&25.5 (25.5 +/- 1.3)&24.1 (24.1 +/- 1.4) p = 0.0020*\\
Dogri&52.1 (52.1 +/- 2.4) p = 0.0010*&NA&53.9 (53.8 +/- 2.3)&52.1 (52.1 +/- 2.3) p = 0.0010*\\
Gitksan&14.3 (14.4 +/- 1.6) p = 0.0010*&NA&25.2 (25.3 +/- 2.5) p = 0.2717&25.7 (25.7 +/- 2.5)\\
Guarani&43.9 (44.0 +/- 1.1)&42.2 (42.3 +/- 1.1) p = 0.0010*&42.8 (42.9 +/- 1.1) p = 0.0020*&42.5 (42.6 +/- 1.1) p = 0.0010*\\
Ilokano&53.5 (53.5 +/- 1.6)&51.8 (51.8 +/- 1.6) p = 0.0010*&52.7 (52.7 +/- 1.8) p = 0.0420*&52.3 (52.3 +/- 1.7) p = 0.0020*\\
Kabuverdianu&69.5 (69.5 +/- 0.8)&67.1 (67.1 +/- 1.0) p = 0.0010*&68.5 (68.5 +/- 0.9) p = 0.0020*&68.6 (68.6 +/- 0.8) p = 0.0010*\\
Kachin&22.7 (22.7 +/- 0.9) p = 0.0010*&NA&25.6 (25.6 +/- 0.8)&23.5 (23.5 +/- 1.9) p = 0.0280*\\
Kalamang&11.5 (11.5 +/- 1.2) p = 0.0010*&20.1 (20.3 +/- 5.4) p = 0.0010*&28.4 (28.8 +/- 7.6) p = 0.0100*&38.4 (38.8 +/- 5.9)\\
Kimbundu&18.6 (18.6 +/- 0.5) p = 0.0010*&NA&23.8 (23.8 +/- 0.9) p = 0.0010*&25.9 (25.9 +/- 0.8)\\
Latgalian&49.9 (49.9 +/- 1.2) p = 0.0959&40.2 (40.2 +/- 1.6) p = 0.0010*&48.5 (48.5 +/- 1.3) p = 0.0010*&50.3 (50.3 +/- 1.0)\\
Minangkabau&55.8 (55.8 +/- 1.0)&44.3 (44.3 +/- 1.4) p = 0.0010*&52.5 (52.5 +/- 1.3) p = 0.0010*&54.7 (54.7 +/- 1.0) p = 0.0010*\\
Mizo&37.0 (37.0 +/- 1.1)&35.1 (35.2 +/- 1.2) p = 0.0010*&35.6 (35.6 +/- 1.2) p = 0.0010*&36.2 (36.2 +/- 1.0) p = 0.0010*\\
Natugu&13.1 (13.1 +/- 0.8) p = 0.0010*&6.4 (6.4 +/- 0.7) p = 0.0010*&9.6 (9.6 +/- 1.2) p = 0.0010*&23.0 (23.0 +/- 2.6)\\
Wolof&37.3 (37.3 +/- 0.9)&30.2 (30.1 +/- 0.9) p = 0.0010*&31.9 (31.9 +/- 1.0) p = 0.0010*&36.7 (36.7 +/- 0.8) p = 0.0050*\\
\bottomrule
\end{tabular}
\caption{\label{tab:bootstrap-ke}
paired bootstrap resampling for chrF++ scores for X-to-English translations. The best performing system is selected as the baseline and is compared to the other systems, using a 95\% confidence interval and a p-value of 0.05. Statistically significant differences are noted with an asterisk (*).}
\end{sidewaystable}

\begin{table*}
\centering
\begin{tabular}{l|c@{ }c@{ }c@{ }c@{ }|c@{ }c@{ }c@{ }c@{ }}
\toprule
 & \multicolumn{4}{c|}{English$\rightarrow$X} & \multicolumn{4}{c}{X$\rightarrow$English} \\
\textbf{Language} & \textbf{Baseline} & \textbf{W} & \textbf{W+S} & \textbf{W+S+G} & \textbf{Baseline} & \textbf{W} & \textbf{W+S} & \textbf{W+S+G} \\
\midrule
Chokwe&0.0&NA&\textbf{1.9}&1.2 &\textbf{6.4}&NA&6.2&5.5 \\
Chuvash&0.3&0.5&\textbf{1.6}&0.7&\textbf{4.3}&1.3&1.8&3.3 \\ 
Dinka&0.0&NA&\textbf{1.5}&0.7&3.5&NA&5.7&\textbf{6.3}\\
Dogri&0.5&NA&\textbf{10.6}&3.2&23.2&NA&\textbf{24.7}&22.6\\
Gitksan&0.0&NA&0.2&\textbf{1.0}&0.2&NA&2.5&\textbf{5.3}\\
Guarani&5.1&1.7&5.3&\textbf{5.6}&\textbf{17.9}&15.5&16.3&16.8\\
Ilokano&14.6&10.8&\textbf{16.1}&15.1&\textbf{28.2}&25.5&26.1&27.0\\
Kabuverdianu&11.0&3.9&\textbf{27.8}&18.1&\textbf{46.5}&41.3&44.0&45.4 \\
Kachin&0.3&NA&\textbf{3.0}&1.9&2.9&NA&\textbf{3.3}&2.5\\
Kalamang&0.0&7.5&\textbf{13.2}&12.2&0.2&2.0&4.4&\textbf{13.9}\\
Kimbundu&0.1&NA&\textbf{4.1}&1.0&0.9&NA&3.0&\textbf{5.0}\\
Latgalian&3.7&1.5&\textbf{10.5}&6.0&21.8&9.7&17.8&\textbf{22.8}\\
Minangkabau&13.0&3.7&\textbf{17.2}&15.8&\textbf{30.0}&12.9&23.2&28.3 \\
Mizo&\textbf{7.6}&6.0&5.4&6.2 &\textbf{10.9}&8.5&8.9&10.0\\
Natugu&0.0&0.0&0.7&\textbf{2.5}&0.1&0.0&0.4&\textbf{5.9}\\
Wolof&3.5&1.1&4.5&\textbf{5.7}&\textbf{12.9}&5.3&6.3&11.4\\
\bottomrule
\end{tabular}
\caption{Collective Table of Results. BLEU scores are shown for all systems. For each of our scores, the combination of reference material that led to the best score is \textbf{bolded}.}
\label{tab:bleu-results}
\end{table*}

\begin{table*}
\centering
\small
\begin{tabular}{p{2cm}|p{8cm}|p{2cm}|p{2cm}}
\toprule
\textbf{Language} & \textbf{Grammar Book} & \textbf{Number of Tokens} & \textbf{Perplexity}\\
\midrule
Chokwe & Martins, João Vicente. (1990) Elementos de Gramática de Utchokwe. Lisboa: Instituto de Investigação Científica Tropical. &	114483 & 23.61 \\
Chuvash& Krueger, John R. (1961) Chuvash Manual: Introduction, Grammar, Reader, and Vocabulary (Indiana University Publications: Uralic and Altaic Series 7). Bloomington: Indiana University. &	118294 & 85.73 \\
Dinka& Nebel, Arturo. (1948) Dinka Grammar (Rek-Malual Dialect) with Texts and Vocabulary. Verona: Istituto Missioni Africane. &	120420 & 55.57 \\
Dogri	& Gupta, Veena. (2014) Dogri. In Omkar N. Koul (ed.), The Languages of Jammu and Kashmir (People's Linguistic Survey of India XII), 3-68. New Delhi: Orient Blackswan. &53993 & 22.38 \\
Gitksan& Hunt, Katharine Dorothy. (1993) Clause Structure, Agreement and Case in Gitksan. University of British Columbia doctoral dissertation. &	106310 & 23.22 \\
Guarani& Gregores, Emma and Jorge A. Suárez. (1967) A Description of Colloquial Guaraní (Janua Linguarum: Series Practica 27). Berlin: Mouton de Gruyter. &	76725 & 19.86 \\
ilokano& Espiritu, Precy. (1984) Let's speak Ilokano. Honolulu: University of Hawaii Press. &	83025 & 26.06 \\
Kabuverdianu& Baptista, Marlyse. (1997) The Morpho-Syntax of Nominal and Verbal Categories in Capeverdean Creole. Harvard University doctoral dissertation. &	104185 & 17.08 \\
Kachin& Hertz, Henry Felix. (1902) A practical handbook of the Kachin or Chingpaw language: containing the grammatical principles and peculiarities of the language, colloquial exercises, and a vocabulary, with an appendix on Kachin customs, laws, and religion. Rangoon: Superintendent of Government Printing, Burma. &	110639 & 33.81 \\
Kalamang& Eline Visser. A grammar of Kalamang. Number 4 in Comprehensive Grammar Library.
Language Science Press, Berlin, 2022. &	92009 & 25.72 \\
Kimbundu & Pedro, José. (1993) Étude grammaticale du kimbundu (Angola). Université de Paris V - René Descartes doctoral dissertation. &119545 & 20.95 \\
Latgalian& Nau, Nicole. (2011) A short grammar of Latgalian (Languages of the World/Materials 482). München: Lincom. &	80567 & 30.71 \\
Minangkabau& Crouch, Sophie. (2009) Voice and verb morphology in Minangkabau, a language of West Sumatra, Indonesia. University of Western Australia MA thesis. &	110746 & 16.05 \\
Mizo& Chhangte, Lalnunthangi. (1993) Mizo Syntax. Eugene: University of Oregon doctoral dissertation. &	85609 & 30.96 \\
Natugu& Boerger, Brenda H. (2022) A Grammar Sketch of Natqgu [ntu]: An Oceanic language of Santa Cruz, Solomon Islands (Texts in the Indigenous Languages of the Pacific 4). Port Moresby: LSPNG. &	80401 & 21.37 \\
Wolof& Ngom, Fallou. (2003) Wolof (Languages of the World/Materials 333). München: Lincom. &	42898 & 11.60 \\
\bottomrule
\end{tabular}
\caption{\label{tab:grammar-book-stats}
Grammar Books and Size}
\end{table*}

\begin{table*}[t]
\centering
\begin{tabular}{lc|l@{}c@{ }c@{ }|c@{ }c@{ }}
\toprule
 & & \multicolumn{3}{c|}{\textbf{Sentences}} & \multicolumn{2}{c}{\textbf{Dictionary Words}} \\
\textbf{Language} & \textbf{ISO 639-3} &\textbf{Source} &\textbf{Train} & \textbf{Test} & \textbf{eng $\rightarrow$ X} & \textbf{X $\rightarrow$ eng} \\
\midrule
Chokwe & cjk & FLORES & 997 & 1012 & 35 & 40\\
Chuvash	& chv & FLORES & 497 & 500 & 3941 & 3611\\
Dinka & dik & FLORES & 997 & 1012 & 10 & 10\\
Dogri & dgo & FLORES & 497 & 500 & 19 & 20\\
Gitksan & git & SIGMORPHON 2023 ST & 42 & 68 & 17 & 16 \\
Guarani & gug & FLORES & 997 & 1012 & 3641 & 3531\\
Ilokano & ilo & FLORES & 997 & 1012 & 5479 & 4779 \\
Kabuverdianu & kea & FLORES & 997 & 1012 & 1413 & 1320 \\
Kachin & kac & FLORES & 997 & 1012 & 92 & 105 \\
Kalamang & kgv & MTOB & 376 & 50 & 1932 & 2531 \\
Kimbundu & kmb & FLORES & 997 & 1012 & 67 & 61 \\
Latgalian & ltg & FLORES & 997 & 1012 & 925 & 710 \\
Minangkabau & min & FLORES & 997 & 1012 & 348 & 349 \\
Mizo & lus & FLORES & 997 & 1012 & 16717 & 14981 \\
Natugu & ntu & SIGMORPHON 2023 ST  & 890 & 99 & 351 & 382 \\
Wolof & wol & FLORES & 997 & 1012 & 2397 & 2850 \\

\bottomrule
\end{tabular}
\caption{Number of sentences and number of words in the dictionaries}
\label{tab:words-sentences-stats}
\end{table*}

\onecolumn
\section{Prompt Format}
\label{app:prompt}

Each sentence to be translated is formatted into a prompt for GPT-4. The prompt has five components: prefix, words, sentences, grammar book, and suffix. The experiment configuration determines whether words (W), sentences (S), or grammar books (G) are included in the prompt. The prefix and suffix are always included in the prompt. In the following sections, we show the format of the prompt by example, using an Ilokano-to-English translation task. We heavily used the code provided by the authors of "Machine Translation from One Book" to generate the prompts. 
\subsection{Prefix}
The prefix provides the task to perform (translation), the source and target languages, and the sentence to translate.
\begin{tcolorbox}
You are an expert translator. Translate the following sentence from Ilokano to English: Adu pay ti babbabassit a klase ti pusa ngem kadakuada a mangmangan iti babbabassit a klase ti ayup a kas iti kuneho, antelope, ken ugsa.
\end{tcolorbox}

\subsection{Words}
For words, we attempt to retrieve the item from the bilingual dictionary. For each word in the source sentence, the top two matching words from the dictionary, as measured by LCS, are included in the prompt.
\begin{tcolorbox}
To help with the translation, here is one of the closest entries to Adu in the bilingual dictionary:\\
Ilokano word: Adams\\
English translation: Adams\\

To help with the translation, here is one of the closest entries to Adu in the bilingual dictionary:\\
Ilokano word: adu\\
English translation: many; lots of; majority; many; much\\

To help with the translation, here is one of the closest entries to pay in the bilingual dictionary:\\
Ilokano word: payso\\
English translation: correct; right\\

To help with the translation, here is one of the closest entries to pay in the bilingual dictionary:\\
Ilokano word: pay\\
English translation: just; please; again; still; yet; also\\

\textit{Additional word-level translations are provided for the remaining words of the source sentence.}
\end{tcolorbox}

\subsection{Sentences}
For sentences, we attempt to retrieve similar samples from our small corpus of parallel sentences. For each word in the source sentence, we find sentences that contain that word, as measured by LCS, and include the top two matches in the prompt.
\begin{tcolorbox}
To help with the translation, here is a translated sentence with words similar to "Adu" in a list of translated reference sentences:\\
Ilokano sentence: Adu dagti restaurant iti aglawlaw ti hardin, ket no iti malem ken rabii masansan nga adda dagiti libre a konsiero iti akintengnga a gazebo.\\
English translation: There are a number of restaurants surrounding the garden, and in the afternoons and evening there free concerts are often given from the central gazebo.\\

To help with the translation, here is a translated sentence with words similar to "Adu" in a list of translated reference sentences:\\
Ilokano sentence: Adu a gobierno ti mangsapul ti bakuna para iti nadumaduma a sakit para kadagiti sangaili a sumrek, wenno dagiti residente a rumuar iti pagilianda.\\
English translation: Many governments require visitors entering, or residents leaving, their countries to be vaccinated for a range of diseases.\\

\textit{Additional sentence-level translations are provided for the remaining words of the source sentence.}
\end{tcolorbox}

\subsection{Grammar Book}
We include the full grammar book in the prompt.
\begin{tcolorbox}{
To help with the translation, here is the full text of a bilingual grammar book:\\
---\\
\#\# FULL BOOK INSERTED HERE \#\#\\
This is the end of the bilingual grammar book.\\
---}
\end{tcolorbox}

\subsection{Suffix}
The suffix reiterates the task and prompts for the appropriate translation.
\begin{tcolorbox}
Now write the translation.\\
Ilokano: Adu pay ti babbabassit a klase ti pusa ngem kadakuada a mangmangan iti babbabassit a klase ti ayup a kas iti kuneho, antelope, ken ugsa.\\
English translation:
\end{tcolorbox}

\end{document}